\documentclass{article}
\usepackage{ijcai20}

\usepackage{times}

\usepackage{soul}
\usepackage{multirow}
\usepackage{makecell}
\usepackage{url}
\usepackage[hidelinks]{hyperref}
\usepackage[utf8]{inputenc}
\usepackage[small]{caption}
\usepackage{graphicx}
\usepackage{amsmath}
\usepackage{booktabs}
\usepackage{bm}
\usepackage{booktabs,threeparttable}
\usepackage{flushend}
\usepackage{color}

\title{Towards Real-Time DNN Inference on Mobile Platforms with Model Pruning and Compiler Optimization}

\author{
Wei Niu$^1$\footnote{These authors contributed equally}\and
Pu Zhao$^2$\footnotemark[1]\and
Zheng Zhan$^2$\And
Xue Lin$^2$\and
Yanzhi Wang$^2$\and 
Bin Ren$^1$\\
\affiliations
$^1$College of William and Mary\\
$^2$Northeastern University\\
\emails
wniu@email.wm.edu,
\{zhao.pu, zhan.zhe\}@husky.neu.edu, \\
\{xue.lin, yanz.wang\}@northeastern.edu,
bren@cs.wm.edu
}

\begin{document}

\maketitle

\begin{abstract}
High-end mobile platforms rapidly serve as primary computing  devices for  a wide range of Deep Neural Network (DNN) applications. However, the constrained computation and storage resources on these devices still pose significant challenges for real-time DNN inference executions. 
To address this problem,
we  propose a set of hardware-friendly structured model pruning and compiler optimization techniques to accelerate DNN executions on mobile devices. 
This demo shows that these optimizations  can enable real-time mobile execution of multiple DNN applications, including style transfer, DNN coloring and super resolution.


\end{abstract}

\section{Introduction}

There are two key phenomena in machine learning and mobile computing fields. First, various Deep Neural Networks (DNN) have served as the fundamental building block of a broad spectrum of machine learning applications because of its superior accuracy and self adaptiveness ability~\cite{goodfellow2016deep}.
Second, with the rapidly increasing popularity of mobile phones, high-end mobile platforms (rather than desktops or servers) serve as primary computing devices, especially for many DNN applications such as wearable devices, video streaming, smart health devices, etc.~\cite{philipp2011sensor,lane2015early}.

It is desirable to deploy real-time DNN inference systems on mobile platforms.
However, due to the intensive computation and high memory storage requirements of state-of-the-art DNN models, such as VGG-16~\cite{simonyan2014very}, ResNet-50~\cite{he2016deep} and MobileNet~\cite{howard2017mobilenets}, it is quite challenging to achieve real-time DNN executions on mobile devices.

Multiple end-to-end mobile DNN acceleration frameworks have been developed, such as TVM~\cite{chen2018tvm}, TensorFlow-Lite (TFLite)~\textcolor{black}{\cite{tensorflowlite}}, and Alibaba Mobile Neural Network (MNN)~\textcolor{black}{\cite{Alibaba}}. However, they still cannot satisfy the real-time execution requirement of DNNs on mobile devices. TVM takes 198 ms to complete the inference of a video frame on an embedded GPU (Adreno 640) with VGG-16, which plays an important role in transfer learning. It takes even longer by TFLite (268 ms). 

This paper investigates DNN inference on mobile platforms with an ultimate goal of real-time execution. The contributions of this paper are summarized as follows:

\begin{itemize}
  \item First, it proposes a set of hardware-friendly structured model pruning and compiler optimization techniques to accelerate DNN executions on mobile devices.
  
  \item Second, it implements and accelerates three interesting and key DNN applications, style transfer~\cite{gatys2016image}, DNN coloring~\cite{2925974}, and super resolution~\cite{dong2014learning} with the help of the proposed model pruning and compiler optimizations.
  
   \item Third, it demonstrates that these three applications can achieve real-time executions on mobile devices. To the best of our knowledge, our implementations are the fastest on mobile devices.
\end{itemize}


\section{Structured Model Pruning}

To satisfy the constraints of computation and storage on mobile devices, 
various DNN model compression techniques  are proposed, among which weight pruning~\cite{luo2017entropy,mao2017exploring,wen2016learning} leads to a notable model size reduction. 

The non-structured weight pruning~\cite{luo2017entropy,guo2016dynamic} is not friendly to modern hardware designs as the indices of the sparse model weights result in stall or complex workload on parallel (specifically, massive parallel) architectures~\cite{wen2016learning}. We mainly explore the more hardware-friendly structured weight pruning~\cite{wen2016learning}, which stores the pruned model regularly in its shape without any weight indices.

Multiple structured pruning approaches exist based on their pruning dimensions, including filter pruning (that prunes the whole filter), channel pruning (that prunes channels), column pruning (that prunes the same location in each filter of each layer), and connectivity and pattern pruning (that prunes both the channels and certain locations in each kernel simultaneously)~\cite{liu2019autoslim,ma2019pconv}. Despite the differences of these structured pruning methods, we support them on a uniform framework based on Alternating Direction Method of Multipliers (ADMM)~\cite{boyd2011distributed}.  In general, the pruning optimization problem is formulated as,
{\small \begin{align}\label{eq: prob}
\min\limits_{\{ \mathbf W _{i}\}}  f( \{\mathbf W _{i} \} ),   \text{subject to} \ \ \mathbf W _{i} \in \mathbf S_i, \forall i
\end{align}}
where $f$ denotes the loss function, $\{\mathbf W _{i} \} $ represents the model weights, and $\mathbf S_i$ shows the  requirement or constraint of the remaining weights in the $i$-th layer to satisfy a certain {\bf structure} (e.g., pre-defined patterns or certain columns/rows preservation). Although the structure constraints make the problem non-differentiable and more complicated, ADMM is able to split the original problem into several easier sub-problems, and iteratively solve them until convergence. 
We apply column pruning for style transfer and kernel pruning for coloring and super resolution.

\section{Compiler Optimization}

Compiler optimizations consist of three components as follows:

\noindent{\bf DSL related optimization}
A DNN model comprises multiple operators (layers) that may show varied computation patterns. A new DSL (i.e. domain specific language) is designed to represent DNN models. This DSL employs a new LR (i.e. layer-wised representation) to represent each layer. Essentially, this DSL is equivalent to the computational graph. Some further computational graph transformation optimizations are also applied to this DSL  (e.g. a combination of Convolution layer/Depthwise Convolution layer + BatchNorm layer + Activation layer) to  reduce the data movement and increase instruction level parallelism.

\noindent{\bf Sparse model storage}
To further improve data locality, the weights of the sparse model are also stored in a more compact format than well-known CSR. This sparse model storage aims to avoid zero-weights storage as CSR with an even better compression ratio by further removing redundant indices generated by our structured pruning. It helps to save the scarce memory-bandwidth of mobile devices.

\noindent{\bf Matrix reorder}
Structured pruning eventually transforms model kernel matrices into small blocks with different pruning patterns. Without any further optimizations, well-know challenges for sparse matrix multiplications still exist, i.e. heavy load imbalance among each thread, and irregular memory accesses. To address this issue, a  matrix  reorder approach is proposed by leveraging the {\bf structure} information offered by our pruning. For example, if a column (and row) pruning is applied. Because this pruning removes all kernel weights in certain  columns  and  rows  within  a  block,  the  remaining weights only appear in other rows and columns with a certain degree of regularity.  Based on this insight, matrix reorder first reorders the rows (e.g., filters in CNN) by arranging the ones with the same or similar patterns together.  Next, it compacts the weights in the column direction (e.g., kernels in CNN).

\section{Experiments and Demonstrations}

This experiment demonstrates the efficacy of the proposed pruning and compilation acceleration approach through three interesting and important DNN applications, style transfer~\cite{gatys2016image}, DNN coloring~\cite{2925974}, and super resolution~\cite{dong2014learning}. The style transfer model is based on a generative network~\cite{zhang2017multistyle} trained on Microsoft COCO~\cite{lin2014microsoft}.  DNN coloring uses the Places scene~\cite{zhou2014learning} dataset to train a novel architecture that can jointly extract  and  fuse global and local features  to perform the final colorization. The super resolution model mainly utilizes residual blocks with wider activation and linear low-rank convolution~\cite{yu2018wide} trained on the DIV2K~\cite{timofte2017ntire} dataset. With structured pruning and compiler optimization, we implement the models on a Samsung Galaxy S10 mobile phone. We demonstrate that our implementations are able to achieve real-time inference on mobile with video demos.

\begin{figure}[tb]    
 \centering
 \scalebox{0.97}[0.97]{
\begin{tabular}{p{0.17in}p{0.8in}p{0.8in}p{0.8in}}
& \parbox{0.9in}{\centering \footnotesize style transfer} &  
\parbox{0.9in}{\centering \footnotesize coloring}
&  
\parbox{0.9in}{\centering \footnotesize super resolution}
\\
 \rotatebox{90}{\parbox{0.9in}{\centering \footnotesize original \\image }}
 &   
\includegraphics[width=0.9in,height=0.9in]{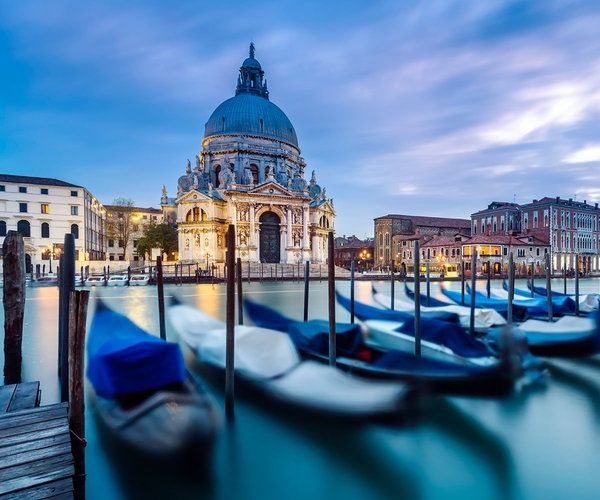}& 
\includegraphics[width=0.9in,height=0.9in]{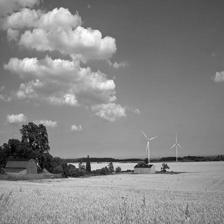}& 
\includegraphics[width=0.9in,height=0.9in]{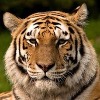}
\\
 \rotatebox{90}{\parbox{0.9in}{\centering \footnotesize application \\ output }}
&   
\includegraphics[width=0.9in,height=0.9in]{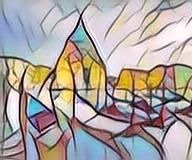} &  
\includegraphics[width=0.9in,height=0.9in]{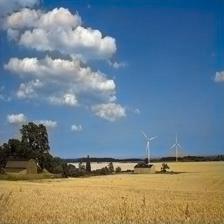} &  
\includegraphics[width=0.9in,height=0.9in]{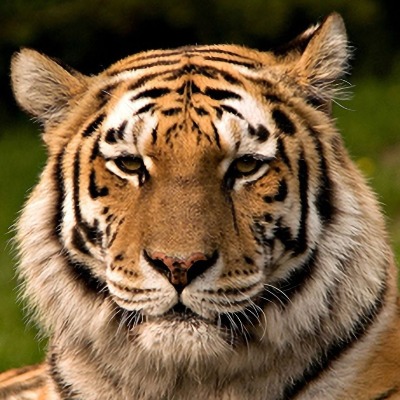}
\end{tabular}}
\caption{\footnotesize{ Examples of  style transfer, coloring, and super resolution implemented on our mobile device.
}} \label{fig: examples}
\end{figure}

Figure \ref{fig: examples} shows sample input and output of three applications.
Table \ref{table_inference} shows the average inference time of these applications on our mobile device. Structured pruning and compiler optimization accelerate the inference with speedups of $4.2\times$, $3.6\times$, and $3.7\times$ for style transfer, coloring and super resolution, respectively. 
 These results demonstrate that our optimized implementation generates satisfied output with high speed on mobile.
 More specifically, all inference can complete within 75 ms, showing the possibility of achieving real-time executions of complex DNN applications on mobile. Please find more video demos at our YouTube channel\footnote{\url{ www.youtube.com/channel/UCCKVDtg2eheRTEuqIJ5cD8A/}.
}.

\begin{table} [tb]
 \centering
  \caption{Average Inference Time on the Mobile Device}
  \label{table_inference}
  \scalebox{0.9}[0.9]{
   \begin{threeparttable}
\begin{tabular}{c|c|c|c}
\toprule[1pt]
 Inference time (ms) & \makecell{Style} &  \makecell{coloring}   &  \makecell{Super resolution} \\ 
\midrule[1pt]
 Unpruned  & 283 &  137 &  269 \\ 
\hline
Pruning  & 178  &  85 & 192 \\ 
\hline
\makecell{ Pruning + compiler } & 67 &  38 & 73 \\ 
\bottomrule[1pt]
 \end{tabular}
\end{threeparttable}}  \vspace*{-6pt}
\end{table}

\section{Potential Impacts}

Real-time DNN executions on mobile have great potential to impact many fields. Take real-time super resolution as an example. It enables users to enjoy high-resolution video streams with limited network bandwidth and inexpensive data cost, while saving providers storage and communication resources. Because only low resolution video streaming is required to store and communicate, super resolution results in significant video service improvement and commercial cost reduction.

\bibliographystyle{named}
\bibliography{ijcai20}




\end{document}